%% file: AIA24_LLM_as_a_Scorer.tex
\documentclass[letterpaper]{article} % DO NOT CHANGE THIS
\usepackage{aaai24}  % DO NOT CHANGE THIS
\usepackage{times}  % DO NOT CHANGE THIS
\usepackage{helvet}  % DO NOT CHANGE THIS
\usepackage{courier}  % DO NOT CHANGE THIS
\usepackage[hyphens]{url}  % DO NOT CHANGE THIS
\usepackage{graphicx} % DO NOT CHANGE THIS
\usepackage{natbib}  % DO NOT CHANGE THIS AND DO NOT ADD ANY OPTIONS TO IT
\usepackage{caption} % DO NOT CHANGE THIS AND DO NOT ADD ANY OPTIONS TO IT
\usepackage{algorithm}
\usepackage{algorithmic}

\usepackage{newfloat}
\usepackage{listings}
\DeclareCaptionStyle{ruled}{labelfont=normalfont,labelsep=colon,strut=off} % DO NOT CHANGE THIS
\floatstyle{ruled}
\newfloat{listing}{tb}{lst}{}
\floatname{listing}{Listing}
\usepackage{subcaption}
\usepackage{microtype}
\usepackage{multirow}
\usepackage{booktabs}

\title{LLM as a Scorer: The Impact of Output Order on Dialogue Evaluation}
\author{
    Yi-Pei Chen\equalcontrib,
    KuanChao Chu\equalcontrib,
    Hideki Nakayama
}
\affiliations{
    The University of Tokyo, Japan\\
    \{ypc, kcchu, nakayama\}@nlab.ci.i.u-tokyo.ac.jp
}

\usepackage{bibentry}
\begin{document}

\maketitle

\begin{abstract}
This research investigates the effect of prompt design on dialogue evaluation using large language models (LLMs). While LLMs are increasingly used for scoring various inputs, creating effective prompts for dialogue evaluation remains challenging due to model sensitivity and subjectivity in dialogue assessments. Our study experimented with different prompt structures, altering the sequence of output instructions and including explanatory reasons. We found that the order of presenting reasons and scores significantly influences LLMs' scoring, with a "reason-first" approach yielding more comprehensive evaluations. This insight is crucial for enhancing the accuracy and consistency of LLM-based evaluations.
\end{abstract}

\input{content}

\section*{Acknowledgements}
This work was supported by the Institute of AI and Beyond of the University of Tokyo and the commissioned research (No. 225) by the National Institute of Information and Communications Technology (NICT), Japan.

\bibliography{aaai24}

\end{document}

%% file: content.tex
\section{Introduction}
% Benefiting from more flexible output formats \cite{instruction_tuning} and alignment with human feedback \cite{ouyang2022training}, Large Language Models (LLMs) \cite{GPT4,llama2} have been increasingly applied to functions requiring human-like subjective judgments, such as scoring.
% Recent advancements in large language models (LLMs) \cite{GPT4,llama2} have led to their increased application in tasks requiring human-like subjective judgments, such as evaluation.
% In the scoring task, a subjective-aware score is given for the input content. 
% Compared to human evaluation, this method achieves significant advantages as for automation and reproducibility \cite{shen2023large}. 
Using large language models (LLMs)~\cite{GPT4,llama2} as evaluators to assign scores to the given inputs have become prevalent. 
\citet{AlphaCode2} output a score between 0 and 1 to estimate the correctness of generated code, thereby ranking their quality. Similarly, \citet{generative_agent} assign poignancy score to the generated text for the retrieval task. 
Other research explores using LLMs to assess generated texts, finding the LLM scores correlates higher with human evaluators than existing automatic metrics \cite{gao2023human,shen2023large,liu-etal-2023-g,luo2023chatgpt}. 
% since they exhibit higher alignment with human judgements and cost lower comparing to human annotators \cite{liu-etal-2023-g,mendonca-etal-2023-simple,shen2023large}.
% Large Language Models (LLMs) \cite{GPT4,llama2} are increasingly being utilized as evaluators to provide subjective scores to input texts.  
% Recent research has shown that LLM evaluators align more closely with human judgment than conventional automated evaluation metrics across various tasks, including dialogue generation \cite{liu-etal-2023-g,mendonca-etal-2023-simple,shen2023large}.

However, designing evaluation prompt for LLMs is not a trivial task, especially for dialogue evaluation. 
Different models exhibit varied sensitivity to the nuances of input prompts. Even slight linguistic variations can lead to significant fluctuations in task performance \cite{leidinger2023linguistic}. 
Moreover, the inherent subjectivity in dialogue evaluation adds on the difficulty and versatility in LLMs' evaluation results. 
While prompt optimization techniques \cite{chen2023instructzero,LLMoptim,zhang2023tempera,prasad2023grips} have been developed to assist in designing more effective prompts, these methods require paired input-output samples for objective value calculation. %as training data. 
Unfortunately, the lack of available dialogue-score pairing data hampers the application of prompt optimization in dialogue evaluation. 
% \citet{zhao2021calibrate} employs a content-free test to adjust output probabilities and mitigate the impact of bias.
% \cite{prasad2023grips} search for optimal prompt modifications with a small portion of samples.
% With a small portion of training I/O samples, \cite{prasad2023grips} enables the search for optimal prompt modifications, \cite{chen2023instructzero} converts the task to a Bayesian optimization problem in the latent space.
% Alternatively, \cite{zhang2023tempera} employs reinforcement learning to automatically obtain prompts optimized for specific tasks. \cite{LLMoptim} utilizes the text interface to simulate LLM as a prompt optimizer.

In this study, we aim to investigate the influence of prompt design on dialogue evaluation, specifically focusing on how the output instructions affects the resulting scores. 
% In this study, we aim to explore the efficacy of prompt design for dialogue evaluation in an interpretable way, particularly focusing on the impact of output format instructions.
%We collect 25 cases of synthesized conversations \footnote{\url{https://reverie.herokuapp.com/arXiv_Demo/}} generated between multiple LLM-powered agents. 
% Our study compares the influence of various output instructions on score outputs using different versions of GPT models, including the synthesis of reasons and the interchange of output instruction order.
We have developed multiple prompt variations to assess the quality of a series of dialogues. %problematic dialogues. %which are known to be problematic. 
These variations involve altering the sequence order of the outputs and examining whether including explanatory reasons along with the scores impacts the evaluation. 
Our analysis compares the influence of different prompts on the scoring outcomes across various versions of GPT models. 
% To this end, we have crafted multiple prompt variations aimed at assessing the quality of a collection of dialogues known to be problematic. Our examination includes analyzing the effects of varying the sequence order of outputs and incorporating explanatory reasons alongside the scores. This research contrasts the impact of different prompts on the scoring outputs across various iterations of GPT models."

We observed that the different order of output instructions can result in different scoring distributions by certain LLMs, even when the corresponding output reasons are similar. %, as shown in Fig.~\ref{fig:tease}. 
Considering the sequential generation nature of auto-regressive models, placing the score after the reasons allows it to reference both the reasons and the input prompt, a dynamic not possible when this order is reversed. %when a score is generated after the reasons, it can reference both the reasons and the input prompt, but not in the opposite. 
The finding suggests that a ``reason-first'' output instruction might lead to a more comprehensive understanding and adherence to the specific requirements of the task. 

\section{Approach}
\input{tables/fig_case}

\input{tables/tab_out_instructs}
% We prepare a task for evaluating a series of dialogues, wherein the LLM will be asked to give an overall score ranging from 1 to 10. This score will represent the collective quality of the dialogues in terms of repetitiveness, factualness, and temporal coherence.
% Given a set of dialogues, the task of LLMs is to rate the dialogues set from 1 to 10, and, if mentioned in the prompt, provide the rational of the rating.
In this study, the task assigned to the LLM is to rate a given set of dialogues on a scale from 1 to 10, where 1 indicates no issues in the set of dialogues, and 10 signifies severe problems. %numerous problems. 
Additionally, if specified in the prompt, the LLM is required to provide a rationale for the rating. 
The dialogues are presented in chronological order, and the output score is determined based on a comprehensive evaluation of the entire set, focusing on key aspects such as repetitiveness, factual accuracy, and coherence.

Along with the task description, we have integrated five customized rules into the prompt, derived from observations in previous experiments without these rules. The special rules include instructions for the LLM to prioritize the number of issues over their impact and to assign more weight to aspects exhibiting significant issues, rather than averaging out the score across all aspects.
% In addition to the task description, we have incorporated five customized rules into the prompt. These special rules are constructed based on our observations from earlier results without them, including instructions for the LLM to prioritize the number of issues over their impact, or to assign greater weight to aspects with significant issues, rather than averaging the evaluation across all aspects, etc.

The final evaluation prompt is organized as follows: a set of dialogues, task description, special rules, and output instruction (see Table~\ref{tab:out_instruct}).
% \begin{itemize}
%     \item A set of dialogues
%     \item Task description
%     \item Special rules
%     \item Output instruction (see Table~\ref{tab:out_instruct})
% \end{itemize}
% \begin{itemize}
%     \item \textbf{%Data}: log of dialogues in chronological order. 
%     \item \textbf{Task description}%instructions regarding the task content.
%     \item \textbf{Special rules} 
%     %For example, instructing the LLM to assign more weights to the aspect that exhibits significant issues rather than averaging all aspects.
%     % \item \textbf{Special rules}: customized scoring policies. 
%     % For instance, instruct the LLM not to calculate an overall score based on the average scores of the three aspects, but rather to give more weight to the one with more serious issues.
%     \item \textbf{Output instruction (Table~\ref{tab:out_instruct})}%: instructions for the output format.
% \end{itemize}
% For each dialogues set, we ran the evaluation  compare the output scores across multiple output instructions for various items and orders.%, as shown in 
For each set of dialogues, we conducted $N$ trials for each of the six configurations (config), varying the output instruction. This experiment was then replicated across $M$ different models.

%as Fig~\ref{fig:method}. 
 
\section{Experiment}
% \subsection{Setting}
\paragraph{Data}
To assess the capability of LLMs in identifying issues within dialogues, we collected LLM-generated dialogues from \citet{generative_agent} and manually grouped them into 25 sets. Each set contains four to six dialogues and exhibits one or more problems, such as repetition or contradictions between dialogues. 
% The ideal outcome for this evaluation would be scores that are closer to 10 instead of 1, reflecting the detection of these issues.
 
% As a result, the desired outcome should be as high as possible. 
% The notation ``case($i,j,k,l,...$)'' refers to a case that contains dialogues with indices ($i,j,k,l,...$). The indices represent the chronological order of the dialogues within the dialogue pool. 
% ``case($i$,$j$,...)'' indicates the dialogues within a case, where $i$ and $j$ are their chronological index in the dialogues pool. 
\paragraph{Model}
We selected four recent LLMs to serve as scorers: gpt-3.5-turbo-0613, gpt-3.5-turbo-1106, gpt-4-turbo-0613 (gpt-4-0613), and gpt-4-1106-preview (gpt-4-1106). 
Note that our aim is to analyze the %variation in 
evaluation scores across various models when altering output instructions, and not to compare them with human judgements for this subjective task.

\input{tables/fig_case_norule}
\subsection{Result and Analysis}

\paragraph{The Importance of Output Instruction Order}
% \paragraph{The order in output instruction matters}

Table~\ref{tab:main} presents the mean scores and standard deviations (std) of 10 trials for all 25 dialogue sets across all configs and models. In both \textit{ex~($\cdot$)} and \textit{json~($\cdot$)} formats, the mean scores for the \textit{rs} settings (output reasons before the score) are generally higher than their \textit{sr} (output score before reasons) counterparts.~\footnote{The exception is observed with the gpt-4-1106 model.}%. 
% \footnote{An exception to this trend is observed with the gpt-4-1106 model, which shows an opposite relationship.}. 
For instance, in the \textit{json~(rs)} config using gpt-4-0613, the mean score is 5.34, while it drops to 3.26 in \textit{json~(sr)}, despite providing similar reasons. 
We conjecture that in the \textit{rs} setting, the autoregressive nature of the model allows the score to be influenced by the previously outputted reasons.
% (TODO, rewrite) One hypothesis is that the sequential generation nature plays a role. In ``sr'' settings, the generated score has no reference to the generated reasons. Conversely, in the 'rs' settings, the generated score can reference on prompt and generated reasons together.

\input{tables/tab_result_main}
%(can we say gpt-4-1106 is less diversed and less creative than gpt-4-0613?) 

\paragraph{Different Levels of Rule Understanding}
In a focused study on a single set with additional 40 trials, as depicted in Fig.~\ref{fig:box_case22}, we observed a trend consistent with the findings presented in Table~\ref{tab:main}. 
However, as shown in Fig.\ref{fig:box_case22_norule}, when we removed the `special rules' from the prompt, we found that most scores were lower and the distinctions between different settings became less pronounced.
This highlights the models' sensitivity to the changes of the prompt.
%This suggests that different configs cause variations in LLM's special rules understanding.

% We study a single case using more trials and plot the score distribution in Fig.~\ref{fig:box_case22}, showing trends similar to those in Table~\ref{tab:main}. In contrast, Fig.~\ref{fig:box_case22_norule} presents the results with ``special rules'' removed. Here, most scores are lower and the difference between settings are less pronounced. This suggests that the difference may related to the understanding of the special rule section. 

%\input{AnonymousSubmission/tables/result_main}
\section{Conclusion}
Our study highlights the sensitivity of LLMs to the order of output instructions, which could be amplified by task-specific rules. 
These findings offer insights for optimizing prompts in subjective tasks like dialogue evaluation.
% These findings illuminate the intricate relationship between prompt design and model output, offering insights for optimizing prompts in subjective evaluation tasks like dialogue evaluation.
% These findings not only demonstrate the nuanced interplay between prompt design and model output but also provide valuable insights for optimizing LLMs in subjective evaluative tasks such as dialogue evaluation.
% We verified that a mere swap in output instructions can lead to different levels of rule understanding and variations in score distribution.
%We verified that a merely swap in output instructions may have different levels of rule understanding and score distribution. %This indicates its key importance in designing prompts for socring.

% \clearpage  

%% file: tables/fig_case.tex
\begin{figure}[t]
\centering

\begin{subfigure}{0.5\columnwidth}
  %\centering
  \includegraphics[width=\linewidth]{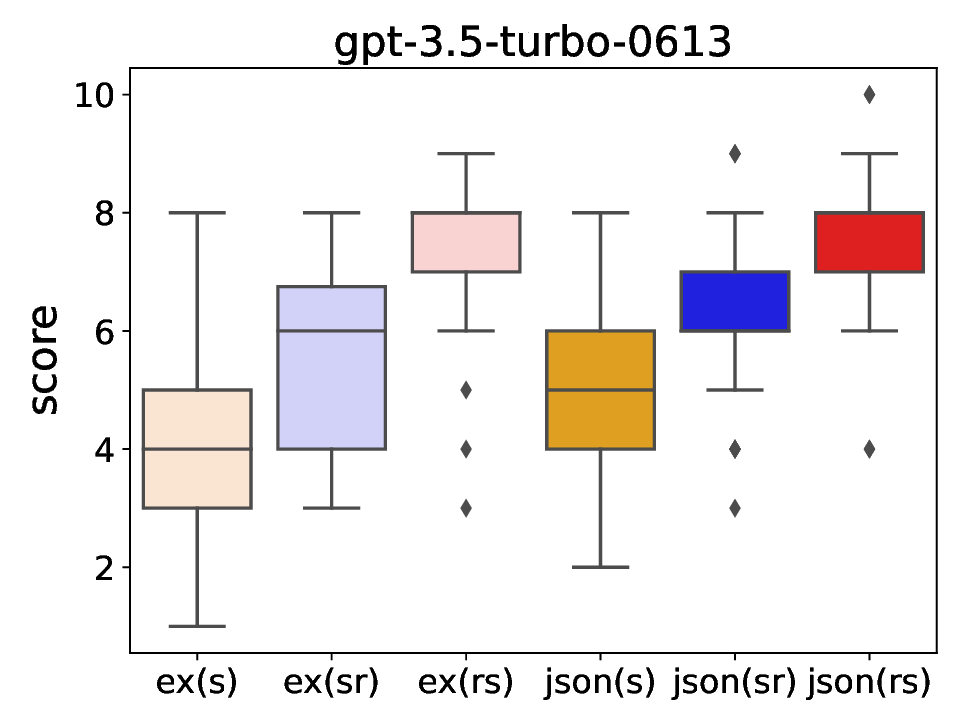}
  %\caption{第一張子圖}
  %\label{fig:sub1}
\end{subfigure}%
\begin{subfigure}{0.5\columnwidth}
  \centering
  \includegraphics[width=\linewidth]{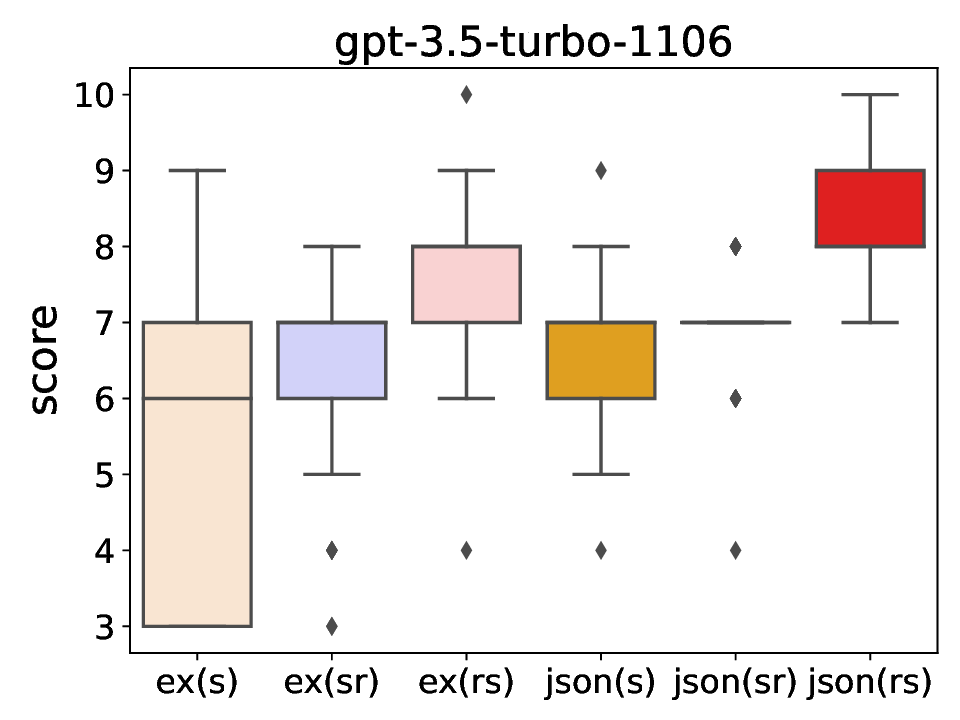}
  %\caption{第二張子圖}
  %\label{fig:sub2}
\end{subfigure}

\begin{subfigure}{0.5\columnwidth}
  \centering
  \includegraphics[width=\linewidth]{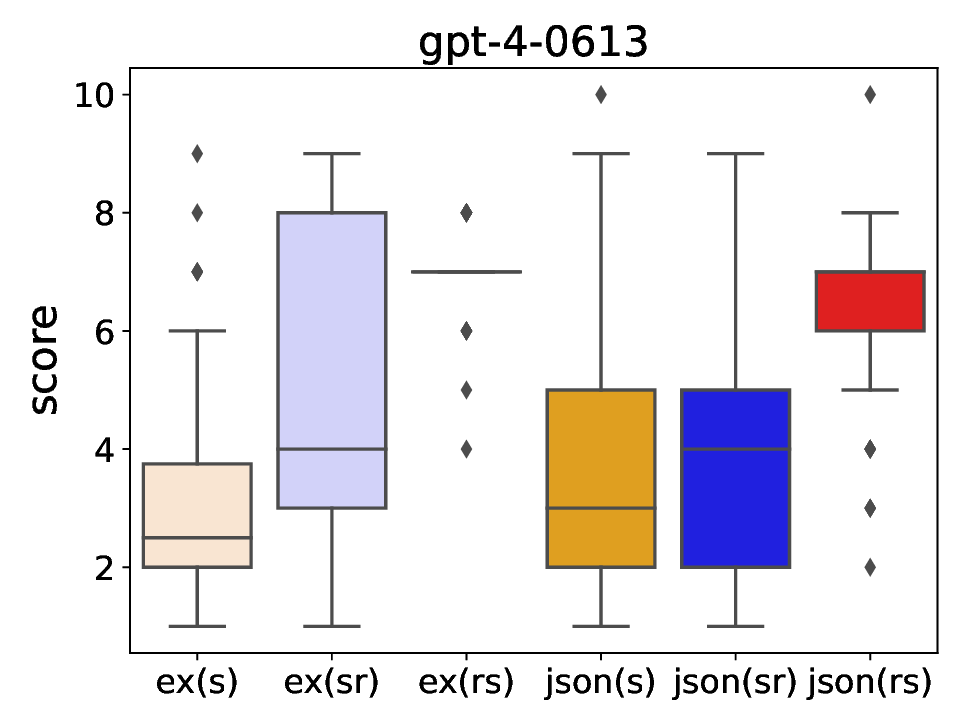}
  %\caption{第三張子圖}
  %\label{fig:sub3}
\end{subfigure}%
\begin{subfigure}{0.5\columnwidth}
  \centering
  \includegraphics[width=\linewidth]{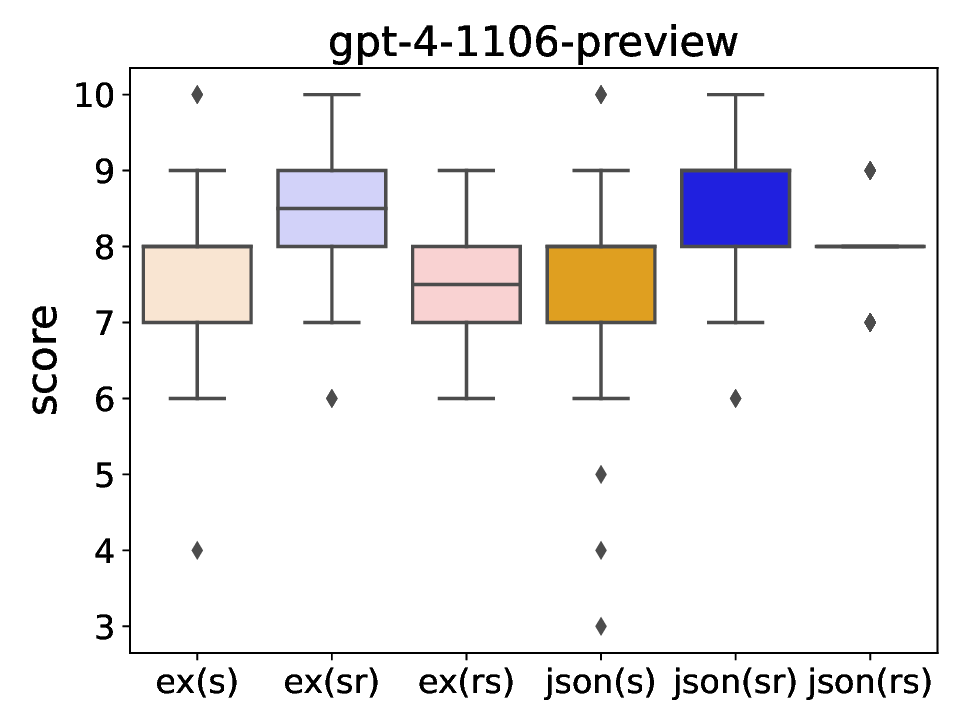}
  %\caption{第四張子圖}
  %\label{fig:sub4}
\end{subfigure}

\caption{Score distribution across 50 trials for each model and output instruction configuration for a dialogue set.}
\label{fig:box_case22}
\end{figure}

%% file: tables/tab_out_instructs.tex
\begin{table}[h]
\centering
\small
%\footnotesize
%\tabcolsep 4pt
\resizebox{1.0\linewidth}{!}{
\begin{tabular}{lp{0.78\linewidth}}
% \hline
\toprule
Config & Output Instruction in the Prompt \\
% \hline
\midrule
ex (s) & \texttt{Example JSON output:}\\
      & \texttt{\{"score": 5\}}\\
% \hline
\midrule          
ex (sr)   & \texttt{Example JSON output:}\\
          & \texttt{\{"score": 5, "reasons": "$<$your reasons for the rating$>$"\}}\\
% \hline
\midrule 
ex (rs)   & swap the order of ``score'' and ``reasons'' in ex (sr)\\
% \hline
\midrule
json (s) & \texttt{Output a json of the following format:} \\
         & \texttt{\{"score": "$<$integer$>$"\}}\\
% \hline
\midrule
json (sr) & \texttt{Output a json of the following format:}\\
          & \texttt{\{"score": "$<$integer$>$", "reasons": "point out the issues and your reasons for the rating"\}} \\
          % & \{\\
          % & \quad``score'': ``$<$integer$>$'',\\
          % & \quad``reasons'': ``point out the issues and your\\ 
          % & \quad\quad\quad\quad\quad\quad  reasons for the rating''\\ % bad
          % & \}\\
% \hline
\midrule
json (rs) & swap the order of ``score'' and ``reasons'' in json (sr)\\
% \hline
\bottomrule
\end{tabular}
}
\caption{The variations of output instruction.}

\label{tab:out_instruct}
\end{table}

%% file: tables/fig_case_norule.tex
\begin{figure}[h]
\centering

\begin{subfigure}{0.5\columnwidth}
  %\centering
  \includegraphics[width=\linewidth]{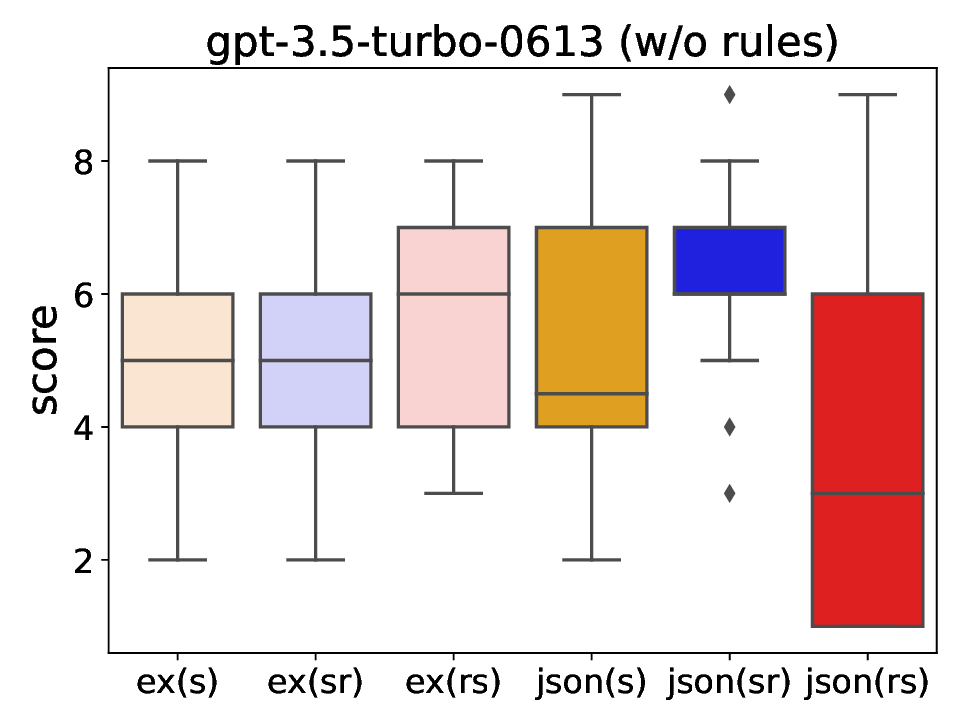}
  %\caption{第一張子圖}
  %\label{fig:sub1}
\end{subfigure}%
\begin{subfigure}{0.5\columnwidth}
  \centering
  \includegraphics[width=\linewidth]{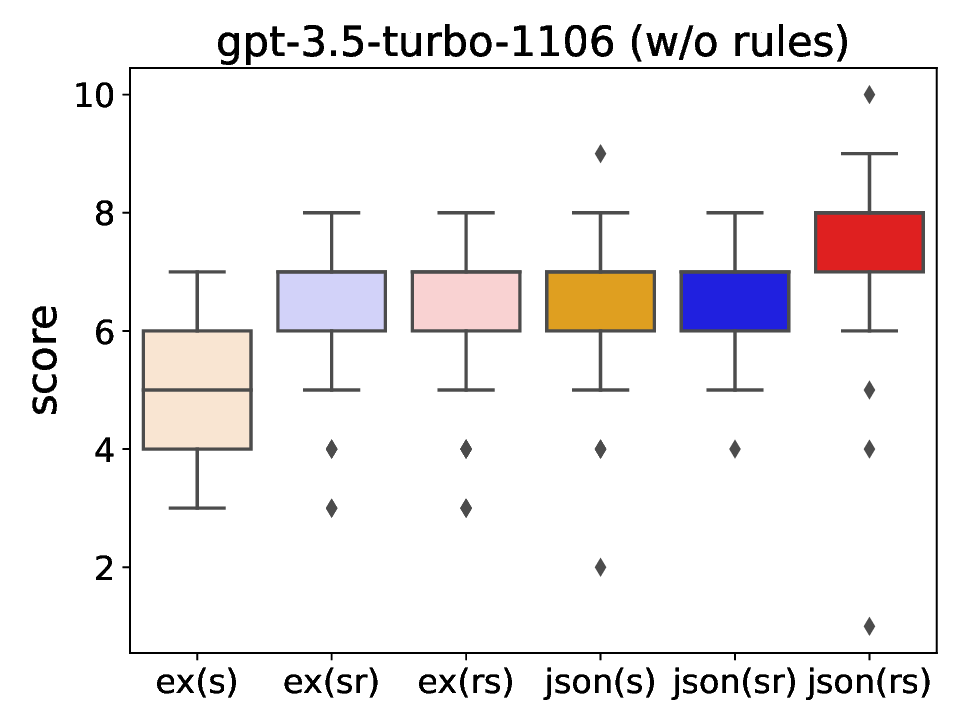}
  %\caption{第二張子圖}
  %\label{fig:sub2}
\end{subfigure}

\begin{subfigure}{0.5\columnwidth}
  \centering
  \includegraphics[width=\linewidth]{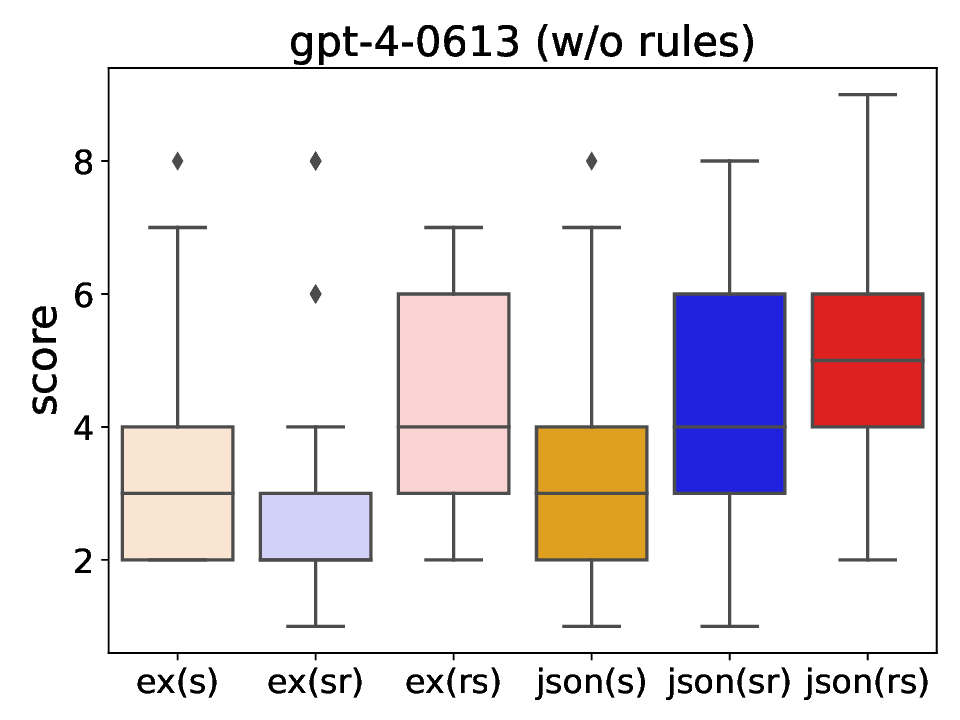}
  %\caption{第三張子圖}
  %\label{fig:sub3}
\end{subfigure}%
\begin{subfigure}{0.5\columnwidth}
  \centering
  \includegraphics[width=\linewidth]{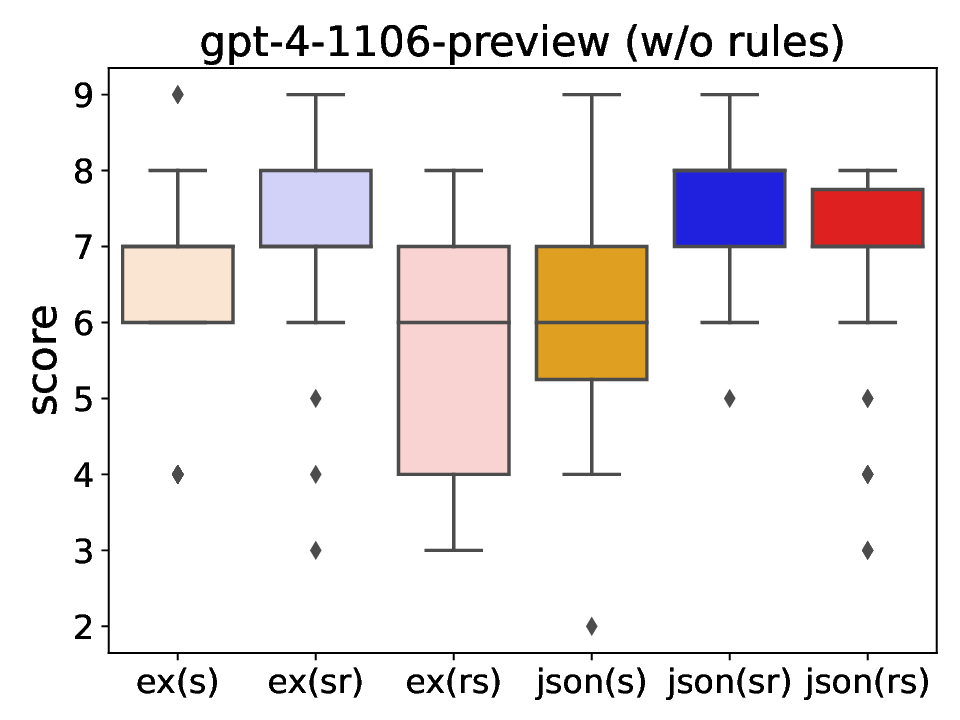}
  %\caption{第四張子圖}
  %\label{fig:sub4}
\end{subfigure}

\caption{
Score distribution across 50 trials for each model and output instruction configuration for a dialogue set, with the `special rules' omitted from the prompt.
% The score distribution from 50 trials for each output instruction and model in case(4,7,9,22), while the ``special rules'' section removed from the prompt.
}
\label{fig:box_case22_norule}
\end{figure}

%% file: tables/tab_result_main.tex
%\begin{table}[]
%    \centering
%    \begin{tabular}{c|c}
%         &  \\
%         & 
%    \end{tabular}
%    \caption{Caption}
%    \label{tab:my_label}
%\end{table}

\begin{table}[h]
\centering
\small
%\footnotesize
\tabcolsep 4pt
\resizebox{1.0\linewidth}{!}{
%Main results. We report the mean (std) of 20 trials for all configurations on each LLM.} %(*GPT-4-1106 is currently a preview version and categorized in ``GPT-4-turbo'' series [?].)}
\begin{tabular}{lcccc}
\toprule
\multirow{2}{*}{Config} &\multicolumn{2}{c}{GPT-3.5-turbo} &\multicolumn{2}{c}{GPT-4}\\
\cmidrule(lr){2-3}\cmidrule(lr){4-5}
  & -0613 & -1106  &-0613 & -1106 \\
\midrule  %$\pm$
ex (s) & 3.68 $\pm$1.17 & 4.51 $\pm$1.19& 3.36 $\pm$1.07 & \textbf{8.18} $\pm$1.05\\

ex (sr) & 4.20 $\pm$1.19 & 5.49 $\pm$1.22 & 3.39 $\pm$1.13 & 7.55 $\pm$1.12\\

ex (rs) & \textbf{6.09} $\pm$1.23 & \textbf{7.66} $\pm$0.81 & \textbf{5.58} $\pm$1.19 & 7.39 $\pm$0.90\\
\midrule
json (s) & 4.03 $\pm$1.16 & 6.18 $\pm$1.09& 3.13 $\pm$1.10 & 6.74 $\pm$1.24\\

json (sr) & 4.66 $\pm$1.15 & 6.76 $\pm$0.94 & 3.26 $\pm$1.11& \textbf{7.69} $\pm$1.06\\

json (rs) & \textbf{5.78} $\pm$1.42 & \textbf{7.99} $\pm$0.94 & \textbf{5.34} $\pm$1.22 & 7.54 $\pm$0.95\\

\bottomrule
\end{tabular}
}
\caption{Mean scores and std for 25 dialogue sets, evaluated across different models and output instruction configurations.}

\label{tab:main}
\end{table}